\documentclass[letterpaper,compsoc,twoside]{IEEEtran}
\usepackage{fixltx2e} 
\usepackage{cmap} 
\usepackage{ifthen}
\usepackage[T1]{fontenc}
\usepackage[utf8]{inputenc}
\usepackage{amsmath}

\usepackage[font={small,it},labelfont=bf]{caption}
\usepackage{float}

\pdfoutput=1
\usepackage{scipy}
\makeatletter
\def\PY@reset{\let\PY@it=\relax \let\PY@bf=\relax%
    \let\PY@ul=\relax \let\PY@tc=\relax%
    \let\PY@bc=\relax \let\PY@ff=\relax}
\def\PY@tok#1{\csname PY@tok@#1\endcsname}
\def\PY@toks#1+{\ifx\relax#1\empty\else%
    \PY@tok{#1}\expandafter\PY@toks\fi}
\def\PY@do#1{\PY@bc{\PY@tc{\PY@ul{%
    \PY@it{\PY@bf{\PY@ff{#1}}}}}}}
\def\PY#1#2{\PY@reset\PY@toks#1+\relax+\PY@do{#2}}

\expandafter\def\csname PY@tok@gd\endcsname{\def\PY@tc##1{\textcolor[rgb]{0.63,0.00,0.00}{##1}}}
\expandafter\def\csname PY@tok@gu\endcsname{\let\PY@bf=\textbf\def\PY@tc##1{\textcolor[rgb]{0.50,0.00,0.50}{##1}}}
\expandafter\def\csname PY@tok@gt\endcsname{\def\PY@tc##1{\textcolor[rgb]{0.00,0.27,0.87}{##1}}}
\expandafter\def\csname PY@tok@gs\endcsname{\let\PY@bf=\textbf}
\expandafter\def\csname PY@tok@gr\endcsname{\def\PY@tc##1{\textcolor[rgb]{1.00,0.00,0.00}{##1}}}
\expandafter\def\csname PY@tok@cm\endcsname{\let\PY@it=\textit\def\PY@tc##1{\textcolor[rgb]{0.25,0.50,0.56}{##1}}}
\expandafter\def\csname PY@tok@vg\endcsname{\def\PY@tc##1{\textcolor[rgb]{0.73,0.38,0.84}{##1}}}
\expandafter\def\csname PY@tok@vi\endcsname{\def\PY@tc##1{\textcolor[rgb]{0.73,0.38,0.84}{##1}}}
\expandafter\def\csname PY@tok@mh\endcsname{\def\PY@tc##1{\textcolor[rgb]{0.13,0.50,0.31}{##1}}}
\expandafter\def\csname PY@tok@cs\endcsname{\def\PY@tc##1{\textcolor[rgb]{0.25,0.50,0.56}{##1}}\def\PY@bc##1{\setlength{\fboxsep}{0pt}\colorbox[rgb]{1.00,0.94,0.94}{\strut ##1}}}
\expandafter\def\csname PY@tok@ge\endcsname{\let\PY@it=\textit}
\expandafter\def\csname PY@tok@vc\endcsname{\def\PY@tc##1{\textcolor[rgb]{0.73,0.38,0.84}{##1}}}
\expandafter\def\csname PY@tok@il\endcsname{\def\PY@tc##1{\textcolor[rgb]{0.13,0.50,0.31}{##1}}}
\expandafter\def\csname PY@tok@go\endcsname{\def\PY@tc##1{\textcolor[rgb]{0.20,0.20,0.20}{##1}}}
\expandafter\def\csname PY@tok@cp\endcsname{\def\PY@tc##1{\textcolor[rgb]{0.00,0.44,0.13}{##1}}}
\expandafter\def\csname PY@tok@gi\endcsname{\def\PY@tc##1{\textcolor[rgb]{0.00,0.63,0.00}{##1}}}
\expandafter\def\csname PY@tok@gh\endcsname{\let\PY@bf=\textbf\def\PY@tc##1{\textcolor[rgb]{0.00,0.00,0.50}{##1}}}
\expandafter\def\csname PY@tok@ni\endcsname{\let\PY@bf=\textbf\def\PY@tc##1{\textcolor[rgb]{0.84,0.33,0.22}{##1}}}
\expandafter\def\csname PY@tok@nl\endcsname{\let\PY@bf=\textbf\def\PY@tc##1{\textcolor[rgb]{0.00,0.13,0.44}{##1}}}
\expandafter\def\csname PY@tok@nn\endcsname{\let\PY@bf=\textbf\def\PY@tc##1{\textcolor[rgb]{0.05,0.52,0.71}{##1}}}
\expandafter\def\csname PY@tok@no\endcsname{\def\PY@tc##1{\textcolor[rgb]{0.38,0.68,0.84}{##1}}}
\expandafter\def\csname PY@tok@na\endcsname{\def\PY@tc##1{\textcolor[rgb]{0.25,0.44,0.63}{##1}}}
\expandafter\def\csname PY@tok@nb\endcsname{\def\PY@tc##1{\textcolor[rgb]{0.00,0.44,0.13}{##1}}}
\expandafter\def\csname PY@tok@nc\endcsname{\let\PY@bf=\textbf\def\PY@tc##1{\textcolor[rgb]{0.05,0.52,0.71}{##1}}}
\expandafter\def\csname PY@tok@nd\endcsname{\let\PY@bf=\textbf\def\PY@tc##1{\textcolor[rgb]{0.33,0.33,0.33}{##1}}}
\expandafter\def\csname PY@tok@ne\endcsname{\def\PY@tc##1{\textcolor[rgb]{0.00,0.44,0.13}{##1}}}
\expandafter\def\csname PY@tok@nf\endcsname{\def\PY@tc##1{\textcolor[rgb]{0.02,0.16,0.49}{##1}}}
\expandafter\def\csname PY@tok@si\endcsname{\let\PY@it=\textit\def\PY@tc##1{\textcolor[rgb]{0.44,0.63,0.82}{##1}}}
\expandafter\def\csname PY@tok@s2\endcsname{\def\PY@tc##1{\textcolor[rgb]{0.25,0.44,0.63}{##1}}}
\expandafter\def\csname PY@tok@nt\endcsname{\let\PY@bf=\textbf\def\PY@tc##1{\textcolor[rgb]{0.02,0.16,0.45}{##1}}}
\expandafter\def\csname PY@tok@nv\endcsname{\def\PY@tc##1{\textcolor[rgb]{0.73,0.38,0.84}{##1}}}
\expandafter\def\csname PY@tok@s1\endcsname{\def\PY@tc##1{\textcolor[rgb]{0.25,0.44,0.63}{##1}}}
\expandafter\def\csname PY@tok@ch\endcsname{\let\PY@it=\textit\def\PY@tc##1{\textcolor[rgb]{0.25,0.50,0.56}{##1}}}
\expandafter\def\csname PY@tok@m\endcsname{\def\PY@tc##1{\textcolor[rgb]{0.13,0.50,0.31}{##1}}}
\expandafter\def\csname PY@tok@gp\endcsname{\let\PY@bf=\textbf\def\PY@tc##1{\textcolor[rgb]{0.78,0.36,0.04}{##1}}}
\expandafter\def\csname PY@tok@sh\endcsname{\def\PY@tc##1{\textcolor[rgb]{0.25,0.44,0.63}{##1}}}
\expandafter\def\csname PY@tok@ow\endcsname{\let\PY@bf=\textbf\def\PY@tc##1{\textcolor[rgb]{0.00,0.44,0.13}{##1}}}
\expandafter\def\csname PY@tok@sx\endcsname{\def\PY@tc##1{\textcolor[rgb]{0.78,0.36,0.04}{##1}}}
\expandafter\def\csname PY@tok@bp\endcsname{\def\PY@tc##1{\textcolor[rgb]{0.00,0.44,0.13}{##1}}}
\expandafter\def\csname PY@tok@c1\endcsname{\let\PY@it=\textit\def\PY@tc##1{\textcolor[rgb]{0.25,0.50,0.56}{##1}}}
\expandafter\def\csname PY@tok@o\endcsname{\def\PY@tc##1{\textcolor[rgb]{0.40,0.40,0.40}{##1}}}
\expandafter\def\csname PY@tok@kc\endcsname{\let\PY@bf=\textbf\def\PY@tc##1{\textcolor[rgb]{0.00,0.44,0.13}{##1}}}
\expandafter\def\csname PY@tok@c\endcsname{\let\PY@it=\textit\def\PY@tc##1{\textcolor[rgb]{0.25,0.50,0.56}{##1}}}
\expandafter\def\csname PY@tok@mf\endcsname{\def\PY@tc##1{\textcolor[rgb]{0.13,0.50,0.31}{##1}}}
\expandafter\def\csname PY@tok@err\endcsname{\def\PY@bc##1{\setlength{\fboxsep}{0pt}\fcolorbox[rgb]{1.00,0.00,0.00}{1,1,1}{\strut ##1}}}
\expandafter\def\csname PY@tok@mb\endcsname{\def\PY@tc##1{\textcolor[rgb]{0.13,0.50,0.31}{##1}}}
\expandafter\def\csname PY@tok@ss\endcsname{\def\PY@tc##1{\textcolor[rgb]{0.32,0.47,0.09}{##1}}}
\expandafter\def\csname PY@tok@sr\endcsname{\def\PY@tc##1{\textcolor[rgb]{0.14,0.33,0.53}{##1}}}
\expandafter\def\csname PY@tok@mo\endcsname{\def\PY@tc##1{\textcolor[rgb]{0.13,0.50,0.31}{##1}}}
\expandafter\def\csname PY@tok@kd\endcsname{\let\PY@bf=\textbf\def\PY@tc##1{\textcolor[rgb]{0.00,0.44,0.13}{##1}}}
\expandafter\def\csname PY@tok@mi\endcsname{\def\PY@tc##1{\textcolor[rgb]{0.13,0.50,0.31}{##1}}}
\expandafter\def\csname PY@tok@kn\endcsname{\let\PY@bf=\textbf\def\PY@tc##1{\textcolor[rgb]{0.00,0.44,0.13}{##1}}}
\expandafter\def\csname PY@tok@cpf\endcsname{\let\PY@it=\textit\def\PY@tc##1{\textcolor[rgb]{0.25,0.50,0.56}{##1}}}
\expandafter\def\csname PY@tok@kr\endcsname{\let\PY@bf=\textbf\def\PY@tc##1{\textcolor[rgb]{0.00,0.44,0.13}{##1}}}
\expandafter\def\csname PY@tok@s\endcsname{\def\PY@tc##1{\textcolor[rgb]{0.25,0.44,0.63}{##1}}}
\expandafter\def\csname PY@tok@kp\endcsname{\def\PY@tc##1{\textcolor[rgb]{0.00,0.44,0.13}{##1}}}
\expandafter\def\csname PY@tok@w\endcsname{\def\PY@tc##1{\textcolor[rgb]{0.73,0.73,0.73}{##1}}}
\expandafter\def\csname PY@tok@kt\endcsname{\def\PY@tc##1{\textcolor[rgb]{0.56,0.13,0.00}{##1}}}
\expandafter\def\csname PY@tok@sc\endcsname{\def\PY@tc##1{\textcolor[rgb]{0.25,0.44,0.63}{##1}}}
\expandafter\def\csname PY@tok@sb\endcsname{\def\PY@tc##1{\textcolor[rgb]{0.25,0.44,0.63}{##1}}}
\expandafter\def\csname PY@tok@k\endcsname{\let\PY@bf=\textbf\def\PY@tc##1{\textcolor[rgb]{0.00,0.44,0.13}{##1}}}
\expandafter\def\csname PY@tok@se\endcsname{\let\PY@bf=\textbf\def\PY@tc##1{\textcolor[rgb]{0.25,0.44,0.63}{##1}}}
\expandafter\def\csname PY@tok@sd\endcsname{\let\PY@it=\textit\def\PY@tc##1{\textcolor[rgb]{0.25,0.44,0.63}{##1}}}

\def\PYZus{\char`\_}

\makeatother

\providecommand*{\DUrole}[2]{%
  \ifcsname DUrole#1\endcsname%
    \csname DUrole#1\endcsname{#2}%
  \else
    \ifcsname docutilsrole#1\endcsname%
      \csname docutilsrole#1\endcsname{#2}%
    \else%
      #2%
    \fi%
  \fi%
}

\ifthenelse{\isundefined{\hypersetup}}{
  \usepackage[colorlinks=true,linkcolor=blue,urlcolor=blue]{hyperref}
  \urlstyle{same} 
}{}

\begin{document}
\newcounter{footnotecounter}\title{Text comparison using word vector representations and dimensionality reduction}\author{Hendrik Heuer$^{\setcounter{footnotecounter}{1}\fnsymbol{footnotecounter}\setcounter{footnotecounter}{2}\fnsymbol{footnotecounter}}$%
          \setcounter{footnotecounter}{1}\thanks{\fnsymbol{footnotecounter} %
          Corresponding author: \protect\href{mailto:hendrikh@kth.se}{hendrikh@kth.se}}\setcounter{footnotecounter}{2}\thanks{\fnsymbol{footnotecounter} Aalto University, Finland}\thanks{%

          \noindent%
          Copyright\,\copyright\,2015 Hendrik Heuer. This is an open-access article distributed under the terms of the Creative Commons Attribution License, which permits unrestricted use, distribution, and reproduction in any medium, provided the original author and source are credited. http://creativecommons.org/licenses/by/3.0/%
        }}\maketitle
          \renewcommand{\leftmark}{PROC. OF THE 8th EUR. CONF. ON PYTHON IN SCIENCE (EUROSCIPY 2015)}
          \renewcommand{\rightmark}{TEXT COMPARISON USING WORD VECTOR REPRESENTATIONS AND DIMENSIONALITY REDUCTION}

\setcounter{page}{13}
\newcommand*{\docutilsroleref}{\ref}
\newcommand*{\docutilsrolelabel}{\label}
\AtEndDocument{\cleardoublepage}
\begin{abstract}This paper describes a technique to compare large text sources using word vector representations (word2vec) and dimensionality reduction (t-SNE) and how it can be implemented using Python. The technique provides a bird’s-eye view of text sources, e.g. text summaries and their source material, and enables users to explore text sources like a geographical map. Word vector representations capture many linguistic properties such as gender, tense, plurality and even semantic concepts like \textquotedbl{}capital city of\textquotedbl{}. Using dimensionality reduction, a 2D map can be computed where semantically similar words are close to each other. The technique uses the word2vec model from the gensim Python library and t-SNE from scikit-learn.\end{abstract}

\section{Introduction%
  \label{introduction}%
}

When summarizing a large text, only a subset of the available topics and stories can be taken into account. The decision which topics to cover is largely editorial. This paper introduces a tool that assists this editorial process using word vector representations and dimensionality reduction. It enables a user to visually identify agreement and disagreement between two text sources.

There are a variety of different ways to approach the problem of visualizing the topics present in a text. The simplest approach is to look at unique words and their occurrences and visualize the words in a list. Topics could also be visualized using word clouds, where the font size of a word is determined by the frequency of the word. Word clouds have a variety of shortcomings: they can only visualize small subsets, they focus on the most frequent words and they do not take synonyms and semantically similar words into account.

This paper describes a human-computer interaction-inspired approach of comparing two text sources. The approach yields a bird’s-eye view of different text sources, including text summaries and their source material, and enables users to explore a text source like a geographical map.
As similar words are close to each other, the user can visually identify clusters of topics that are present in the text.

This paper describes a tool, which can be used to visualize the topics in a single text source as well as to compare different text sources. To compare the topics in source A and source B, three different sets of words can be computed: a set of unique words in source A, a set of unique words in source B as well as the intersection set of words both in source A and B. These three sets are then plotted at the same time. For this, a colour is assigned to each set of words. This enables the user to visually compare the different text sources and makes it possible to see which topics are covered where. The user can explore the word map and zoom in and out. He or she can also toggle the visibility, i.e. show and hide, certain word sets.

The comparison can be used to visualize the difference between a text summary and its source material. It can also help to compare Wikipedia revisions in regards to the topics they cover. Another possible application is the visualization of heterogeneous data sources like a list of search queries and keywords.

The Github repository of the tool includes an online demo {[}Heu15{]}. The tool can be used to explore the precomputed topic sets of the Game of Thrones Wikipedia article revisions from 2013 and 2015. The repository also includes the precomputed topic sets for the Wikipedia article revisions for the articles on World War 2, Facebook, and the United States of America.

\section{Distributional semantic models%
  \label{distributional-semantic-models}%
}

The distributional hypothesis by Harris states that words with similar meaning occur in similar contexts {[}Sah05{]}. This implies that the meaning of a word can be inferred from its distribution across contexts. The goal of distributional semantics is to find a representation, e.g. a vector, that approximates the meaning of a word {[}Bru14{]}. The traditional approach to statistical modeling of language is based on counting frequencies of occurrences of short word sequences of length up to N and did not exploit distributed representations {[}Cun15{]}.  Distributional semantics takes word co-occurrence in context windows into account.

The general idea behind word space models is to use distributional statistics to generate high-dimensional vector spaces, where a word is represented by a context vector that encodes semantic similarity {[}Sah05{]}. The representations are called distributed representations because the features are not mutually exclusive and because their configurations correspond to the variations seen in the observed data {[}Cun15{]}. LeCun et al. provide the example of a news story. When the task is to predict the next word in a news story, the learned word vectors for Tuesday and Wednesday will be very similar as they can be easily replaced by each other when used in a sentence {[}Cun15{]}.

There are a variety of computational models that implement the distributional hypothesis, including word2vec {[}Che13{]}, GloVe {[}Pen14{]}, dependency-based word embeddings {[}Lev14{]} and Random Indexing {[}Sah05{]}. There are a variety of Python implementations of these techniques. word2vec is available in gensim {[}Řeh10{]}. For GloVe, the C source code was ported to Python {[}Gau15, Kul15{]}. The dependency-based word embeddings by Levy and Goldberg are implemented in spaCy {[}Hon15{]}. Random Indexing is available in an implementation by Joseph Turian {[}Tur10{]}.

For this paper, word2vec was selected because Mikolov et al. provide 1.4 million pre-trained entity vectors trained on 100 billion words from various news articles in the Google News dataset {[}Che13{]}. However, other models might perform equally well for the purpose of text comparison. Moreover, custom word vectors trained on a large domain-specific dataset, e.g. the Wikipedia encyclopedia for the Wikipedia revision comparison, could potentially yield even better results.

\subsection{word2vec%
  \label{word2vec}%
}

word2vec is a tool developed by Mikolov, Sutskever, Chen, Corrado, and Dean at Google. The two model architectures in the C tool were made available under an open-source license {[}Mik13{]}. Gensim provides a Python reimplementation of word2vec {[}Řeh10{]}.

Word vectors encode semantic meaning and capture many different degrees of similarity {[}Lev14{]}. word2vec word vectors can capture linguistic properties such as gender, tense, plurality, and even semantic concepts such as \textquotedbl{}is the capital city of\textquotedbl{}. word2vec captures domain similarity while other more dependency-based approaches capture functional similarity.

In the word2vec vector space, linear algebra can be used to exploit the encoded dimensions of similarity. Using this, a computer system can complete tasks like the Scholastic Assessment Test (SAT) analogy quizzes, that measure relational similarity.\begin{equation*}
king - man + woman \approx queen
\end{equation*}It works for the superlative:\begin{equation*}
fastest - fast + slow \approx slowest
\end{equation*}As well as the past participle:\begin{equation*}
woken - wake + be \approx been
\end{equation*}It can infer the Finnish national sport from the German national sport.\begin{equation*}
football - Germany + Finland \approx hockey
\end{equation*}Based on the last name of the current Prime Minister of the United Kingdom, it identifies the last name of the German Bundeskanzlerin:\begin{equation*}
Cameron - England + Germany \approx Merkel
\end{equation*}The analogies can also be applied to the national dish of a country:\begin{equation*}
haggis - Scotland + Germany \approx Currywurst
\end{equation*}Fig. 1 shows the clusters of semantically similar words and how they form semantic units, which can be easily interpreted by humans.\begin{figure}[]\noindent\makebox[\columnwidth][c]{\includegraphics[width=\columnwidth]{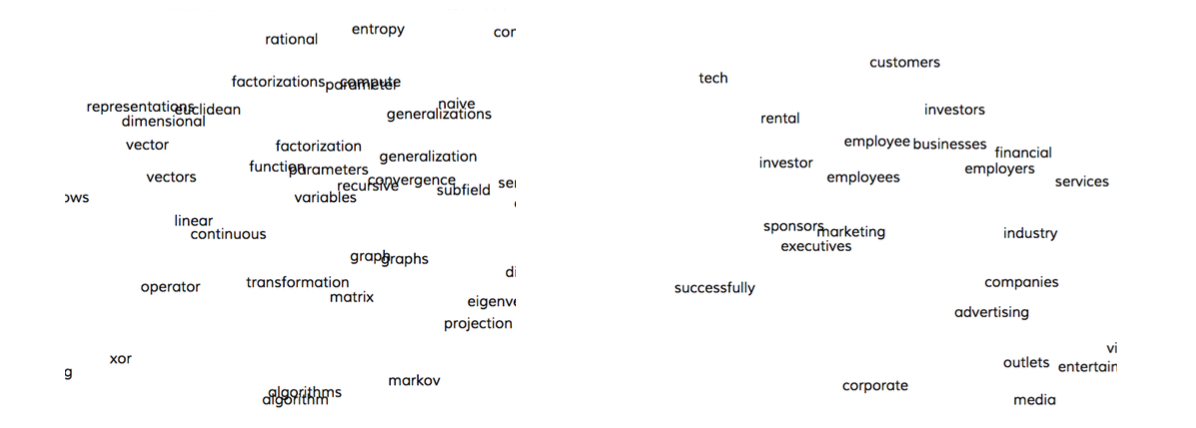}}
\caption{Clusters of semantically similar words emerge when the word2vec vectors are projected down to 2D using t-SNE. \DUrole{label}{egfig}}
\end{figure}

\section{Dimensionality reduction with t-SNE%
  \label{dimensionality-reduction-with-t-sne}%
}

t-distributed Stochastic Neighbour Embedding (t-SNE) is a dimensionality reduction technique that retains the local structure of data and that helps to visualize large real-world datasets with limited computational demands {[}Maa08{]}. Vectors that are similar in a high-dimensional vector space get represented by two- or three-dimensional vectors that are close to each other in the two- or three-dimensional vector space. Dissimilar high-dimensional vectors are distant in the two- or three-dimensional vector space. Meanwhile, the global structure of the data and the presence of clusters at several scales is revealed. t-SNE is well-suited for high-dimensional data that lies on several different, but related, low-dimensional manifolds {[}Maa08{]}.

t-SNE achieves this by minimizing the Kullback-Leibler divergence between the joint probabilities of the high-dimensional data and the low-dimensional representation. The Kullback-Leibler divergence measures the faithfulness with which a probability distribution q represents a probability distribution p by a discrete scalar and equals zero if the distributions are the same {[}Maa08{]}. The Kullback-Leibler divergence is minimized using the gradient descent method. In contrast to other Stochastic Neighbor Embedding methods that use Gaussian distributions, it uses a Student t-distribution.

\section{Implementation%
  \label{implementation}%
}

The text comparison tool implements a workflow that consists of a Python tool for the back-end and a Javascript tool for the front-end. With the Python tool, a text is converted into a collection of two-dimensional word vectors. These are visualized using the Javascript front-end. With the Javascript front-end, the user can explore the word map and zoom in and out to investigate both the local and the global structure of the text sources. The Javascript front-end can be published online.

The workflow of the tool includes the following four steps:

\subsection{Pre-processing%
  \label{pre-processing}%
}

In the pre-processing step, all sentences are tokenized to extract single words. The tokenization is done using the Penn Treebank Tokenizer implemented in the Natural Language Processing Toolkit (NLTK) for Python {[}Bir09{]}. Alternatively, this could also be achieved with a regular expression.

Using a hash map, all words are counted. Only unique words, i.e. the keys of the hash map, are taken into account for the dimensionality reduction. The 3000 most frequent English words according to a frequency list collected from Wikipedia are ignored to reduce the amount of data.

\subsection{Word representations%
  \label{word-representations}%
}

For all unique non-frequent words, the word representation vectors are collected from the word2vec model from the gensim Python library {[}Řeh10{]}. Each word is represented by an N-dimensional vector (N=300, informed by the best accuracy in {[}Mik13{]} and following the default in {[}Che13{]}).\begin{Verbatim}[commandchars=\\\{\},fontsize=\footnotesize]
\PY{k+kn}{from} \PY{n+nn}{gensim.models} \PY{k+kn}{import} \PY{n}{Word2Vec}

\PY{n}{model} \PY{o}{=} \PY{n}{Word2Vec}\PY{o}{.}\PY{n}{load\PYZus{}word2vec\PYZus{}format}\PY{p}{(}
 \PY{n}{word\PYZus{}vectors\PYZus{}filename}\PY{p}{,} \PY{n}{binary}\PY{o}{=}\PY{n+nb+bp}{True}
\PY{p}{)}

\PY{k}{for} \PY{n}{word} \PY{o+ow}{in} \PY{n}{words}\PY{p}{:}
  \PY{k}{if} \PY{n}{word} \PY{o+ow}{in} \PY{n}{model}\PY{p}{:}
    \PY{k}{print} \PY{n}{model}\PY{p}{[}\PY{n}{word}\PY{p}{]}
\end{Verbatim}

\subsection{Dimensionality Reduction%
  \label{dimensionality-reduction}%
}
The resulting N-dimensional word2vec vectors are projected down to 2D using the t-SNE Python implementation in scikit-learn {[}Ped11{]}.

In the dimensionality reduction step, the N-dimensional word vectors are projected down to a two-dimensional space so that they can be easily visualized in a 2D coordinate system (see Fig. 2).\begin{figure}[]\noindent\makebox[\columnwidth][c]{\includegraphics[width=\columnwidth]{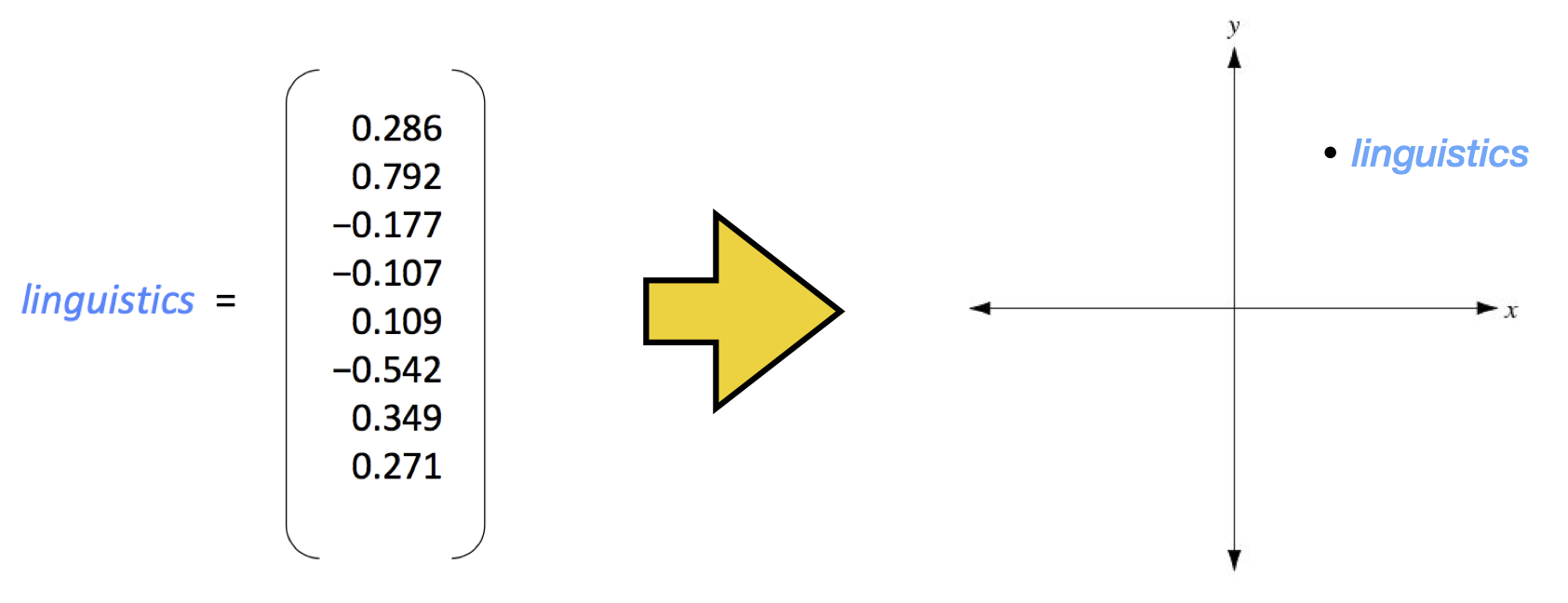}}
\caption{In the dimensionality reduction step, the word vectors are projected down to 2D. \DUrole{label}{egfig}}
\end{figure}

For the implementation, the t-SNE implementation in scikit-learn is used:\begin{Verbatim}[commandchars=\\\{\},fontsize=\footnotesize]
\PY{k+kn}{from} \PY{n+nn}{sklearn.manifold} \PY{k+kn}{import} \PY{n}{TSNE}

\PY{n}{tsne} \PY{o}{=} \PY{n}{TSNE}\PY{p}{(}\PY{n}{n\PYZus{}components}\PY{o}{=}\PY{l+m+mi}{2}\PY{p}{)}
\PY{n}{tsne}\PY{o}{.}\PY{n}{fit\PYZus{}transform}\PY{p}{(}\PY{n}{word\PYZus{}vectors}\PY{p}{)}
\end{Verbatim}

\subsection{Visualization%
  \label{visualization}%
}
After the dimensionality reduction, the vectors are exported to a JSON file. The vectors are visualized using the D3.js JavaScript data visualization library {[}Bos12{]}. Using D3.js, an interactive map was developed. With this map, the user can move around and zoom in and out. The colour coding helps to judge the ratio of dissimilar and similar words. At the global scale, the map can be used to assess how similar two text sources are to each other. At the local scale, clusters of similar words can be explored.

\section{Results%
  \label{results}%
}

As with many unsupervised methods, the evaluation can be difficult and the quality of the visualization is hard to quantify. The goal of this section is, therefore, to introduce relevant use cases and illustrate how the technique can be applied.

The flow described in the previous section can be applied to different revisions of Wikipedia articles. For this, a convenience sample of the most popular articles in 2013 from the English Wikipedia was used.  For each article, the last revision from the 31st of December 2013 and the most recent revision on the 26th of May 2015 were collected. The assumption was that popular articles will attract sufficient changes to be interesting to compare. The list of the most popular Wikipedia articles includes Facebook, Game of Thrones, the United States, and World War 2.

The article on Game of Thrones was deemed especially illustrative for the task of comparing the topics in a text, as the storyline of the TV show developed between the two different snapshot dates as new characters were introduced. Other characters became less relevant and were removed from the article. The article on World War 2 was especially interesting as one of the motivations for the topic tool is to find subtle changes in data.

Fig. 3 shows how different the global cluster, i.e. the full group of words on the minimum zoom setting, of the Wikipedia articles on the United States, Game of Thrones and World War 2 are.\begin{figure}[]\noindent\makebox[\columnwidth][c]{\includegraphics[width=\columnwidth]{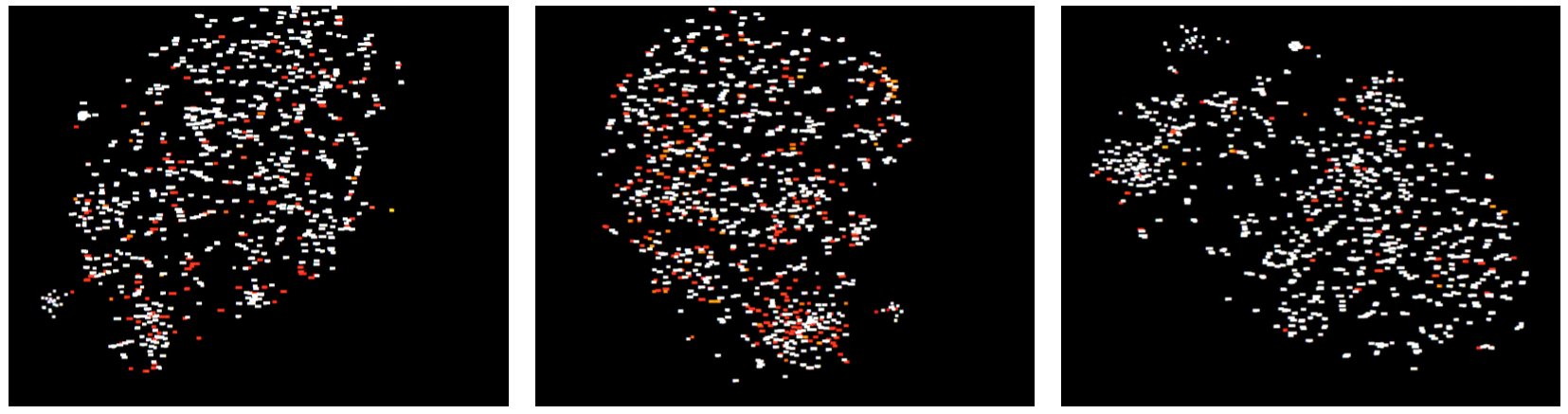}}
\caption{Global clusters of the Wikipedia articles on the United States (left), Game of Thrones (middle), and World War 2 (right). \DUrole{label}{egfig}}
\end{figure}

Fig. 4 shows four screenshots of the visualization of the Wikipedia articles on the United States including an overview and detail views that only show the intersection set of words, words only present in the 2013 revision of the article, and words only present in the 2015 revision of the article.\begin{figure}[]\noindent\makebox[\columnwidth][c]{\includegraphics[width=\columnwidth]{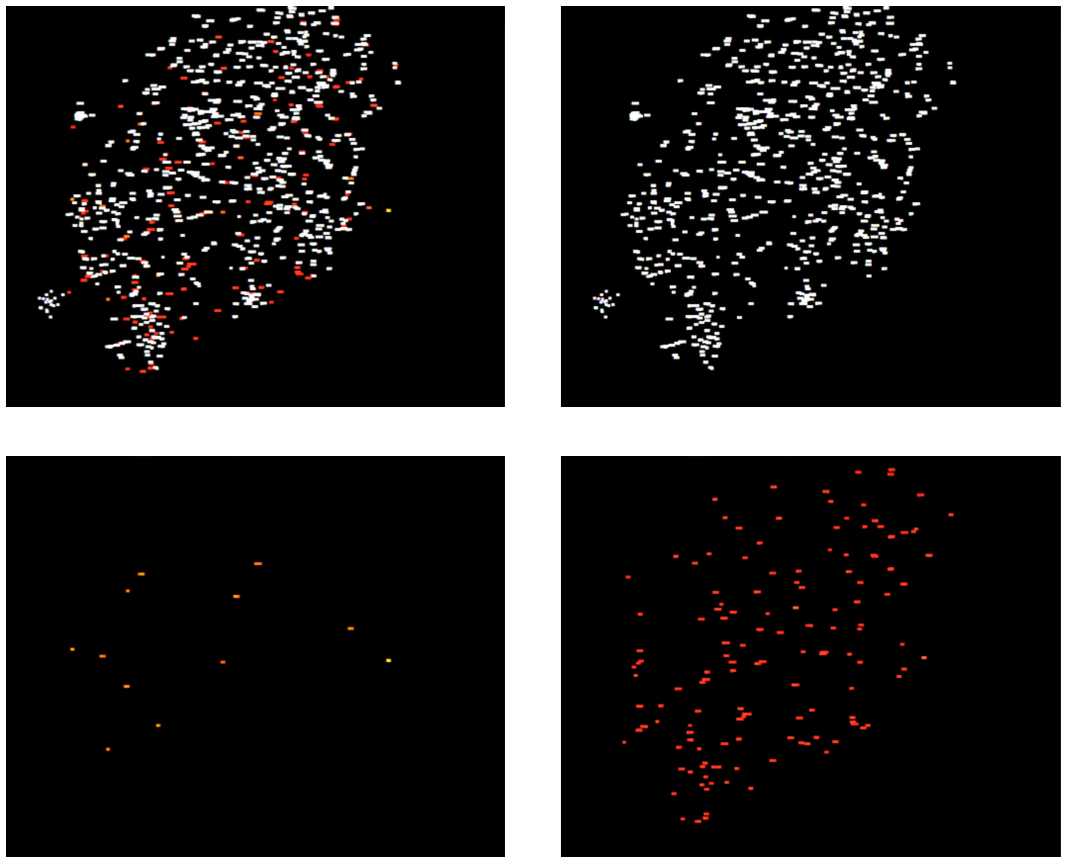}}
\caption{Topic comparison of the Wikipedia article on the United States. In the top left, all words in both texts are plotted. On the top right, only the intersection set of words is shown. In the bottom left, only the words present in the 2013 revision are displayed. In the bottom right, only the words present in the 2015 revision are shown. \DUrole{label}{egfig}}
\end{figure}

When applied to Game of Thrones, it is e.g. easy to visually compare names that were removed since 2013 and that were added in 2015 (Fig. 5). Using the online demo available {[}Heu15{]}, this technique can be applied to the Wikipedia articles on the United States and World War 2.

The technique can also be applied to visualize the Google search history of an individual. Similar words are represented by similar vectors. Thus, terms related to different topics, e.g. technology, philosophy or music, will end up in separate clusters.\begin{figure}[]\noindent\makebox[\columnwidth][c]{\includegraphics[width=\columnwidth]{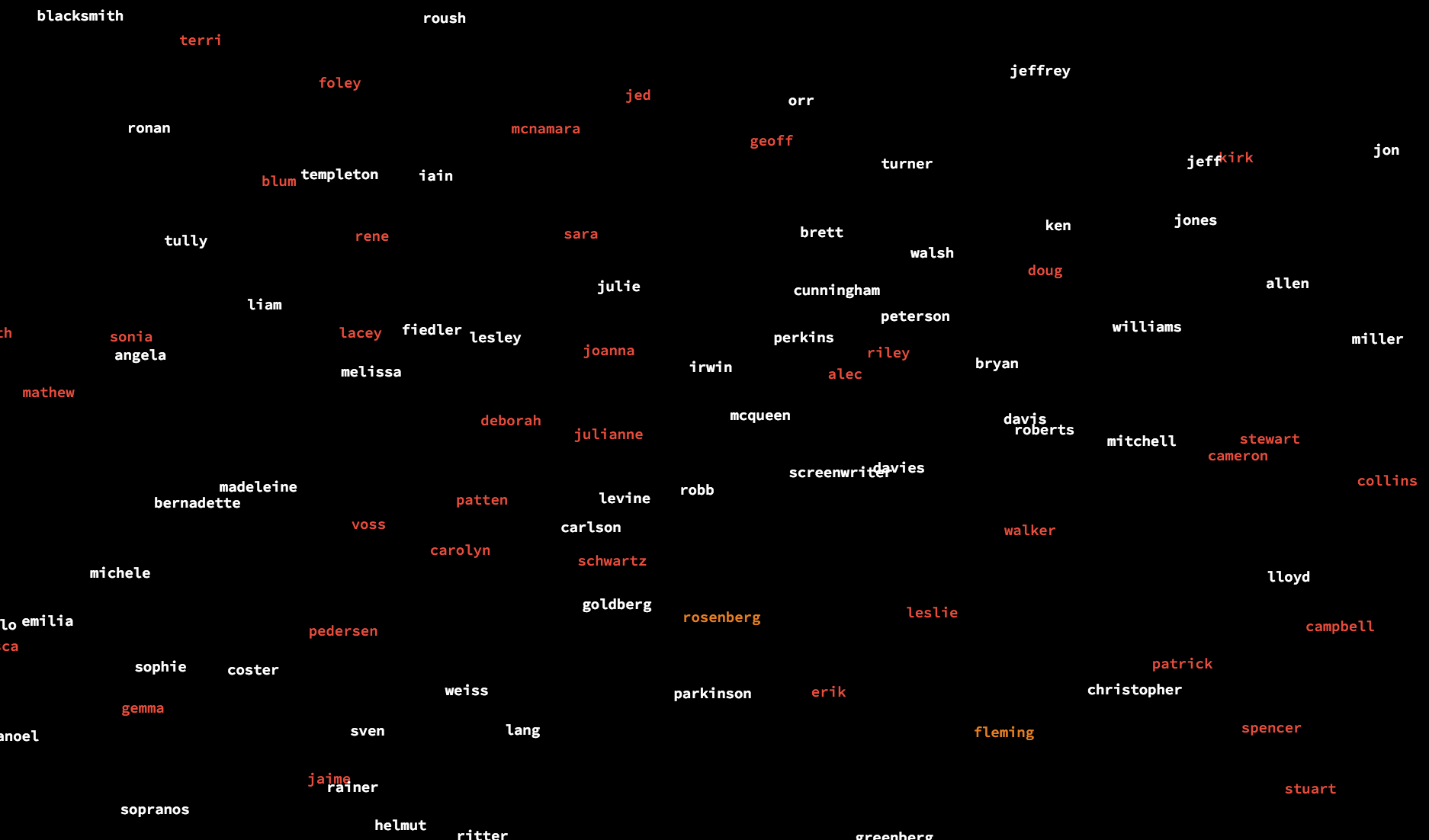}}
\caption{Names present in the Wikipedia article on Game of Thrones. Red names were added to the 2015 revision, orange names removed. White names are present in both revisions.}
\end{figure}

\section{Conclusion%
  \label{conclusion}%
}

Word2vec word vector representations and t-SNE dimensionality reduction can be used to provide a bird’s-eye view of different text sources, including text summaries and their source material. This enables users to explore a text source like a geographical map.

The paper gives an overview of an ongoing investigation of the usefulness of word vector representations and dimensionality reduction in the text and topic comparison context. The major flaw of this paper is that the introduced text visualization and text comparison approach is not validated empirically.

As many researchers publish their source code under open source licenses and as the Python community embraces and supports these publications, it was possible to integrate the findings from the literature review of my Master's thesis into a useable tool. Distributed representations are an active field of research. New findings on word, sentence or paragraph vectors can be easily integrated into the workflow of the tool.

Both the front-end and the back-end of the implementation were made available on GitHub under GNU General Public License 3 {[}Heu15{]}. The repository includes the necessary Python code to collect the word2vec representations using Gensim, to project them down to 2D using t-SNE and to output them as JSON. The repository also includes the front-end code to explore the JSON file as a geographical map.

The tool can be used in addition to topic modeling techniques like LDA. It enables the comparison of large text sources at a glance and is aimed at similar text sources with subtle differences.

\end{document}